\documentclass[letterpaper]{article} 
\usepackage{aaai25}  
\usepackage{times}  
\usepackage{helvet}  
\usepackage{courier}  
\usepackage[hyphens]{url}  
\usepackage{graphicx} 
\usepackage{natbib}  
\usepackage{caption} 
\usepackage{algorithm}
\usepackage{algorithmic}

\usepackage{newfloat}
\usepackage{listings}
\DeclareCaptionStyle{ruled}{labelfont=normalfont,labelsep=colon,strut=off} 
\floatstyle{ruled}
\newfloat{listing}{tb}{lst}{}
\floatname{listing}{Listing}
\usepackage{booktabs}
\usepackage{multirow}
\usepackage{array}
\usepackage{amsmath}
\usepackage{amssymb}

\title{Cherry-Picking in Time Series Forecasting: How to Select Datasets to Make Your Model Shine}
\author {
    Luis Roque\textsuperscript{\rm 1, \rm 2},
    Carlos Soares\textsuperscript{\rm 1, \rm 2, \rm 3},
    Vitor Cerqueira\textsuperscript{\rm 1, \rm 2},
    Luis Torgo\textsuperscript{\rm 4}
}
\affiliations {
    \textsuperscript{\rm 1}Faculdade de Engenharia, Universidade do Porto, Porto, Portugal\\
    \textsuperscript{\rm 2}Laboratory for AI and CS (LIACC), Porto, Portugal\\
    \textsuperscript{\rm 3}Fraunhofer Portugal AICOS, Porto, Portugal\\
    \textsuperscript{\rm 4}Dalhousie University, Halifax, Canada\\
    luis\_roque@live.com
}

\begin{document}

\maketitle

\begin{abstract}
The importance of time series forecasting drives continuous research and the development of new approaches to tackle this problem. Typically, these methods are introduced through empirical studies that frequently claim superior accuracy for the proposed approaches. Nevertheless, concerns are rising about the reliability and generalizability of these results due to limitations in experimental setups. This paper addresses a critical limitation: the number and representativeness of the datasets used.
We investigate the impact of dataset selection bias, particularly the practice of cherry-picking datasets, on the performance evaluation of forecasting methods. Through empirical analysis with a diverse set of benchmark datasets, our findings reveal that cherry-picking datasets can significantly distort the perceived performance of methods, often exaggerating their effectiveness.
Furthermore, our results demonstrate that by selectively choosing just four datasets — what most studies report — 46\% of methods could be deemed best in class, and 77\% could rank within the top three. Additionally, recent deep learning-based approaches show high sensitivity to dataset selection, whereas classical methods exhibit greater robustness. Finally, our results indicate that, when empirically validating forecasting algorithms on a subset of the benchmarks, increasing the number of datasets tested from 3 to 6 reduces the risk of incorrectly identifying an algorithm as the best one by approximately 40\%.
Our study highlights the critical need for comprehensive evaluation frameworks that more accurately reflect real-world scenarios. Adopting such frameworks will ensure the development of robust and reliable forecasting methods.
\end{abstract}

%

\section{Introduction}\label{intro}

Time series forecasting is critical in various application domains, including finance, meteorology, and industry. Over the past decades, there has been significant interest in developing accurate forecasting models, leading to a variety of methods, from traditional statistical techniques to advanced deep learning models.

The selection of datasets for evaluating forecasting models can significantly impact the experimental results. For various reasons, such as reducing computational complexity, researchers often select:

\begin{enumerate}
    \item A limited number of datasets,
    \item Datasets that may not be representative of real-world data,
    \item A subset of time series when working with large datasets, and
    \item A small set of baseline and state-of-the-art (SOTA) models for comparison, often with inconsistent and unfair tuning efforts.
\end{enumerate}

Regarding point 3), recent work addresses this problem and identifies several flaws in the most commonly used datasets in the area of time series anomaly detection. It suggests that comparisons in many papers introducing new approaches might not generalize to the real world \cite{Keogh_anomaly_det}. An example of point 4) is the comparison between simple one-layer linear models and sophisticated Transformer-based time series forecasting models \cite{zeng2022transformerseffectivetimeseries}. To the best of our knowledge, no work has yet been published that attempts to understand the implications of point 1) and 2). In this paper, we focus on understanding the consequences of point 1) and how such selection can introduce bias, impacting the quality and generalizability of the results. 


In the context of dataset selection, we use the term \emph{cherry-picking} for the deliberate or random process of selecting a limited number of datasets that may not be representative of the broader data landscape. This practice involves selecting specific datasets that might showcase the strengths of a model while ignoring others that could reveal its weaknesses. Cherry-picking can lead to biased results and overly optimistic model performance estimates. Thus, it can also significantly impact the quality and generalizability of new forecasting models, making them less reliable in real-world applications.

Our results show that \emph{cherry-picking} specific datasets can significantly distort perceived model performance, even as the number of datasets used for reporting results increases. Our analysis shows that with a commonly used selection of 4 selected datasets, 46\% of models could be reported as the best, and 77\% could be presented within the top 3 positions, highlighting the potential for biased reporting.


The rest of this paper is organized as follows: Section 2 provides background information, including definitions of the forecasting problem and the modeling approaches used. Section 3 describes the materials and methods employed in our empirical analysis. The experiments and results are presented in Section 4 and discussed in Section 5. Finally, we conclude the paper in Section 6.

All experiments are fully reproducible, and the methods and data are available in a public code repository.\protect\footnotemark{}

\footnotetext{\url{https://github.com/luisroque/bench}}

\section{Background}
\label{sec:background}

This section provides an overview of topics related to our work. We begin by defining the problem of time series forecasting from both classical and machine learning perspectives. Next, we discuss the limitations of current evaluation frameworks and highlight recent works that address common problems and inconsistencies. The following two sections review prior work on classical methods and deep learning approaches. Finally, we discuss the evaluation metrics and dataset selection used in forecasting problems.

\subsection{Time Series Forecasting}\label{sec:2.1}

A univariate time series can be represented as a sequence of values \( Y = \{y_1, y_2, \dots, y_t\} \), where \( y_i \in \mathbb{R} \) denotes the value at the \( i \)-th timestep, and \( t \) represents the length of the series. The objective in univariate time series forecasting is to predict future values \( y_{t+1}, \ldots, y_{t+h} \), where \( h \) is the forecasting horizon.

In the context of machine learning, forecasting problems are treated as supervised learning tasks. The dataset is constructed using time delay embedding~\cite{bontempi2013machine}, a technique that reconstructs a time series into Euclidean space by applying sliding windows. This process results in a dataset \(\mathcal{D} = \{\langle X_{i}, y_{i} \rangle\}_{i=p+1}^t\), where \( y_i \) denotes the \( i \)-th observation and \( X_i \in \mathbb{R}^p \) is the corresponding set of \( p \) lags: \( X_i = \{y_{i-1}, y_{i-2}, \dots, y_{i-p}\} \). Time series databases often comprise multiple univariate time series. 

We define a time series database as \( \mathcal{Y} = \{Y_1, Y_2, \dots, Y_n\} \), where \( n \) is the number of time series in the collection. Forecasting methods in these contexts are categorized into local and global approaches \cite{januschowski2020criteria}. Traditional forecasting techniques typically adopt a local approach, wherein an independent model is applied to each time series in the database. Conversely, global methods involve training a single model using all time series in the database, a strategy that has demonstrated superior forecasting performance \cite{godahewa2021ensembles}. This performance improvement is attributed to the fact that related time series within a database—such as the demand series of different related retail products—can share useful patterns. Global models can capture these patterns across different series, whereas local models can only learn dependencies within individual series.

The training of global forecasting models involves combining the data from various time series during the data preparation stage. The training dataset \(\mathcal{D}\) for a global model is a concatenation of individual datasets: \(\mathcal{D} = \{\mathcal{D}_1, \dots, \mathcal{D}_n\}\), where \(\mathcal{D}_j\) represents the dataset corresponding to the time series \(Y_j\). As previously described, the auto-regressive formulation is applied to the combined dataset to facilitate the learning process.

\subsection{Limitations to Current Evaluation Frameworks}\label{sec:2.2}

Recent work has critically evaluated the effectiveness of various experimental setups and how they provided inconsistent results compared to previous works.

An example is the widespread adoption of Transformer-based approaches in time series forecasting, which have consistently outperformed benchmarks. Nonetheless, a recent study raised doubts about the reliability of these results \cite{zeng2022transformerseffectivetimeseries}. It argues that the permutation-invariant self-attention mechanism in Transformers can result in temporal information loss, making these models less effective for time series tasks. The study compares SOTA Transformer-based models with a simple one-layer linear model, which surprisingly outperforms the more complex counterparts across multiple datasets. This suggests that simpler approaches may often be more suitable.

Another critical perspective is offered regarding the limitations of anomaly detection tasks \cite{Keogh_anomaly_det}. In most cases, benchmarks often suffer from issues like triviality, unrealistic anomaly density, and mislabeled ground truth. These flaws can lead to misleading conclusions about the effectiveness of proposed models.

Additional works show that inflated accuracy gains often result from unfair comparisons, such as inconsistent network architectures and embedding dimensions. Also, unreliable metrics and test set feedback further aggravate the issue \cite{musgrave2020metriclearningrealitycheck}. Similarly, many studies report significant improvements over weak baselines without exceeding prior benchmarks \cite{trecretrievalresults}. These findings emphasize the need for stricter experimental rigor and transparent longitudinal comparisons. It is the only way to ensure the reliability of the reported progress.

One study introduces a framework designed to assess the robustness of hierarchical time series forecasting models under various conditions \cite{roque2024rhiots}. Despite the deep learning adoption in the field and their capacity to handle complex patterns, the authors demonstrate that traditional statistical methods often show greater robustness. This happens even in cases when the data distribution undergoes significant changes.


\subsection{Classical Methods}\label{sec:2.3}

Several approaches have been developed to address time series forecasting. Simple methods, such as Seasonal Naive (\texttt{SNaive}), predict future values based on the last observed value from the same season in previous cycles. Classical forecasting methods, including \texttt{ARIMA}, exponential smoothing, and their variations, are favored for their simplicity, interpretability, and robustness \cite{hyndman2018forecasting, gardner1985exponential}. 

\texttt{ARIMA} models, which combine autoregression, differencing, and moving averages, are effective for linear time series with trends and seasonal components. Exponential smoothing methods, such as Holt-Winters, model seasonality and trends through weighted averages.

Nevertheless, these classical methods have limitations. They often require significant manual tuning and assumptions about the underlying data structure. For instance, \texttt{ARIMA} requires stationary data and appropriate differencing parameters, while exponential smoothing methods may struggle with complex seasonal patterns and large datasets.

\subsection{Deep Learning Methods}\label{sec:2.4}

Deep learning models have been showing steady progress in time series forecasting. The initial approach was based on Recurrent Neural Networks (\texttt{RNNs}) \cite{ELMAN1990179}, including Long-Short-Term Memory (\texttt{LSTM}) networks and Gated Recurrent Units (\texttt{GRUs}), which are designed to capture long-term dependencies in sequential data. Nevertheless, they can suffer from issues like vanishing gradients, which can impede their ability to model long sequences effectively.

Then, Convolutional models were adapted to time series, for example, the Temporal Convolutional Networks (\texttt{TCNs}) \cite{lea2016temporalconvolutionalnetworksunified}. They address some of these issues by enabling parallel processing of sequences and capturing long-range dependencies more efficiently.

Recently, Transformer models, initially developed for natural language processing, have been increasingly applied to time series forecasting and have shown better performance than \texttt{RNNs} \cite{zhou2021informerefficienttransformerlong}. Transformers use a self-attention mechanism that allows each part of the input sequence to attend to every other part directly. By avoiding the recurrent structure of \texttt{RNNs}, Transformers can handle long sequences and complex dependencies more effectively. Nevertheless, the self-attention mechanism has limitations due to its quadratic computation and memory consumption on long sequences. The  \texttt{Informer} model was introduced to overcome these computational constraints. From the paper, we see an improvement in accuracy between 1.5 to 2 times the results obtained by an \texttt{LSTM} approach \cite{zhou2021informerefficienttransformerlong}.

Despite the seemingly impressive results from Transformer models, recent studies have shown that simple linear models can outperform Transformers on forecasting benchmarks \cite{zeng2022transformerseffectivetimeseries}. This highlights the potential bias introduced by experimental setups and has renewed interest in simpler and more efficient approaches, such as the \texttt{NHITS} and \texttt{TiDE} models \cite{challu2023nhits,das2024longtermforecastingtidetimeseries}.

The \texttt{NHITS} and \texttt{TiDE} models both utilize Multi-layer Perceptrons (MLPs) to achieve efficient time-series forecasting. \texttt{NHITS} incorporates hierarchical interpolation and multi-rate data sampling techniques, assembling predictions sequentially to emphasize components with different frequencies and scales. This method allows \texttt{NHITS} to efficiently decompose the input signal and synthesize the forecast, making it particularly effective for long-horizon forecasting. Experiments show that \texttt{NHITS} outperforms state-of-the-art methods, improving accuracy by nearly 20\% over recent Transformer models (e.g., \texttt{Informer}) and significantly reducing computation time by an order of magnitude. On the other hand, \texttt{TiDE} is an encoder-decoder model that leverages the simplicity and speed of linear models while handling covariates and non-linear dependencies. The \texttt{TiDE} model claims to surpass \texttt{NHITS} in performance while being 5 to 10 times faster than the best Transformer-based models.

\subsection{Evaluation Metrics}\label{sec:2.6}

Evaluating forecasting performance involves various metrics, which can be scale-dependent, scale-independent, percentage-based, or relative. Common metrics include Mean Absolute Error (MAE), Root Mean Squared Error (RMSE), and symmetric mean absolute percentage error (SMAPE). Hewamalage et al. \cite{hewamalage2023forecast} provide a comprehensive survey of these metrics, offering recommendations for their use in different scenarios.

In the M4 competition \cite{makridakis2018m4}, SMAPE and MASE (Mean Absolute Scaled Error) were used for evaluation:

\begin{equation}
    \text{SMAPE} = \frac{100\%}{n} \sum_{i=1}^{n} \frac{|\hat{y}_i - y_i|}{(|\hat{y}_i| + |y_i|)/2}
\end{equation}

\noindent where $\hat{y}_i$ and $y_i$ are the forecast and actual values for the $i$-th instance, $n$ is the number of observations, and $m$ is the seasonal period. 

\subsection{Dataset Selection in Experimental Evaluations}\label{sec:2.5}

The selection of datasets is a key factor in determining the generalizability and reproducibility of time series forecasting experiments. It directly influences the robustness of the conclusions drawn from experimental results, making it essential for researchers to carefully consider both the type and number of datasets used.

Across the models discussed in this section, the number of datasets used in experimental setups is relatively small, typically ranging from three to six. For instance, \texttt{DeepAR} \cite{salinas2019deeparprobabilisticforecastingautoregressive} uses three standard public datasets: \textit{Parts} \cite{SNYDER2012485}, \textit{Electricity} \cite{misc_electricityloaddiagrams20112014_321}, and \textit{Traffic} \cite{OLIVARES2024470}.

The selection of datasets often reflects the specific goals of each model. For example, models like \texttt{Informer} \cite{zhou2021informerefficienttransformerlong}, \texttt{NHITS} \cite{challu2023nhits}, and \texttt{TiDE} \cite{das2024longtermforecastingtidetimeseries} focus on long-term time series forecasting. They are evaluated using datasets similar to those used by \texttt{DeepAR}, such as \textit{Electricity} and \textit{Traffic}, as well as others like \textit{Weather} \cite{zeng2022transformerseffectivetimeseries}. Additionally, these models utilize the more recently introduced \textit{ETT} series, which was made available by the authors of \texttt{Informer} when releasing their paper. These newer datasets feature a small number of time series but a very large number of observations per series.

It is important to note that \texttt{NHITS}, which evolved from \texttt{N-BEATS} \cite{oreshkin2020nbeatsneuralbasisexpansion}, exclusively adopts a long-term forecasting evaluation setup. In contrast, N-BEATS was originally tested using a more classical forecasting setup with datasets like \textit{Tourism} \cite{athanasopoulos2011tourism}, \textit{M3} \cite{makridakis2000m3}, and \textit{M4} \cite{makridakis2018m4}. These classical datasets are characterized by a significantly larger number of time series, though each series has relatively few observations.

Additionally, models like \texttt{TiDE} \cite{das2024longtermforecastingtidetimeseries} separate their experimental setup into different tasks, differentiating between long-term prediction and demand prediction tasks. For the latter, it uses the \textit{M5} \cite{makridakis2022m5} dataset and compares its accuracy against models like \texttt{DeepAR}.

\begin{table*}[t]
\centering
\caption{Summary of the datasets used in the experimental setup, including the number of time series, number of observations, forecast horizon, and frequency. Sources: Labour \cite{pmlr-v139-rangapuram21a}, M3 \cite{makridakis2000m3}, M4 \cite{makridakis2018m4}, M5 \cite{MAKRIDAKIS20221346}, Tourism \cite{athanasopoulos2011tourism}, Traffic \cite{OLIVARES2024470}, Wiki2 \cite{pmlr-v139-rangapuram21a}, ETTh1, ETTh2 \cite{zhou2021informerefficienttransformerlong}.}
\label{tab:data}
\begin{tabular}{llr@{\hskip 0.3cm}r@{\hskip 0.3cm}r@{\hskip 0.3cm}r@{\hskip 0.3cm}}
\toprule
& & \# time series & \# observations & H & Frequency \\
\midrule
\multirow[t]{2}{*}{Labour} & Monthly & 57 & 28671 & 6 & 12 \\
\midrule
\multirow[t]{3}{*}{M3} & Monthly & 1428 & 167562 & 18 & 12 \\
& Quarterly & 756 & 37004 & 8 & 4 \\
& Yearly & 645 & 18319 & 6 & 1 \\
\midrule
\multirow[t]{3}{*}{M4} & Monthly & 48000 & 11246411 & 18 & 12 \\
& Quarterly & 24000 & 2406108 & 8 & 4 \\
& Yearly & 23000 & 858458 & 6 & 1 \\
\midrule
M5 & Daily & 30490 & 47649940 & 30 & 365 \\
\midrule
\multirow[t]{2}{*}{Tourism} & Monthly & 366 & 109280 & 18 & 12 \\
& Quarterly & 427 & 42544 & 8 & 4 \\
\midrule
Traffic & Daily & 207 & 75762 & 30 & 365 \\
\midrule
Wiki2 & Daily & 199 & 72834 & 30 & 365 \\
\midrule
\multirow[t]{1}{*}{ETTh1} & Hourly & 1 & 17420 & 48 & 24 \\
\midrule
\multirow[t]{1}{*}{ETTh2} & Hourly & 1 & 17420 & 48 & 24 \\
\midrule
Total & & 129577 & 62747833 & - & - \\
\bottomrule
\end{tabular}
\end{table*}

\section{Framework for Evaluating Cherry-Picking}\label{sec:approach}

In this section, we present our framework for assessing cherry-picking in time series forecasting evaluations. Our methodology is designed to systematically evaluate how the selection of specific datasets can bias the reported performance of forecasting models, potentially leading to misleading conclusions.

Cherry-picking refers to the practice of selectively presenting data that supports a desired conclusion while ignoring data that may contradict it. In the context of time series forecasting, this could mean reporting model performance only on datasets where a particular model performs well while omitting cases where it does not. Consider a scenario where you have five different forecasting models and ten datasets, each with unique characteristics like seasonality and trend. If you selectively report the performance of these models on just the datasets where your preferred model performs best, you might claim it as the "top-performing model." Nevertheless, this claim could be misleading if, on the full set of datasets, the model does not perform as well overall. Our framework helps identify whether such cherry-picking has occurred by analyzing the performance of each model across various subsets of the datasets and comparing it to their overall performance.

Our framework involves three key steps: 1) dataset selection and categorization, 2) model selection, 3) performance evaluation and ranking, and 4) empirical analysis. 

Step 1) in our framework is to compile a comprehensive set of benchmark datasets, denoted as \(\mathcal{D} = \{D_1, D_2, \dots, D_m\}\), where each \(D_i\) represents a unique dataset. These datasets should be chosen to cover a wide range of domains, frequencies, and characteristics, such as seasonality, trend, noise, and intermittency. This diversity ensures that the experimental setup can effectively capture different challenges encountered in time series forecasting.

In step 2), we select a diverse set of forecasting models, denoted as \(\mathcal{M} = \{M_1, M_2, \dots, M_n\}\), where each \(M_i\) represents a forecasting model. The models are chosen to represent a broad spectrum of approaches, including both classical methods (e.g., \texttt{ARIMA}, \texttt{ETS}) and advanced deep learning models (e.g., \texttt{Informer}, \texttt{NHITS}, \texttt{TiDE}). This diversity ensures that the analysis captures the performance of both simple statistical models and complex neural networks.

Step 3) involves the performance evaluation and ranking. We evaluate the performance of each model on different subsets of the available datasets. 
For each model \(M_i \in \mathcal{M}\) and each subset \(\mathcal{D}_j \subseteq \mathcal{D}\) of size \(n\), we define the ranking function \(R(M_i, \mathcal{D}_j)\). It assigns a rank to model \(M_i\) based on its SMAPE values across the dataset subset \(\mathcal{D}_j\) where \(|\mathcal{D}_j| = n\). Here, \(n\) represents the specific size of the subsets \(\mathcal{D}_j\) considered from the overall dataset \(\mathcal{D}\), with \(n\) ranging from 1 to \(N\). The models are ranked from 1 to \(m\) (where \(m\) is the total number of models), with rank 1 indicating the best performance (i.e., the lowest SMAPE). This ranking allows us to assess how the relative performance of models changes as the selection of datasets varies.

To assess the impact of cherry-picking, in step 4), we perform the following empirical analysis:
\begin{itemize}
    \item \textbf{Baseline Ranking}: Evaluate the performance of each model \(M_i\) on the entire dataset collection \(\mathcal{D}\), establishing a baseline ranking \(R(M_i, \mathcal{D})\), where \(R(M_i, \mathcal{D})\) denotes the rank of model \(M_i\) when evaluated on the full dataset collection \(\mathcal{D}\).
    
    \item \textbf{Top-\( k \) Datasets}: For each model \( M_i \), identify the dataset subsets \( \mathcal{D}_{k}(M_i) \) where the model consistently ranks in the top \( k \). This is done by evaluating the rank \( R(M_i, \mathcal{D}_j) \) for each subset \( \mathcal{D}_j \subseteq \mathcal{D} \) of size \( n \), and selecting the subsets where \( M_i \) achieves one of its top \( k \) ranks.

    \item \textbf{Rank Consistency}: Finally, we evaluate ranking changes as subset size \( n \) increases. We gradually increase the size \( n \) of the dataset subset \( \mathcal{D}_j \) from 1 to \( N \) and observe how the ranking \( R(M_i, \mathcal{D}_j) \) changes as more datasets are included.

\end{itemize}

\section{Experimental Setup}\label{sec:experimental}

This experimental setup illustrates how our framework can be applied to assess the robustness of time series forecasting models. We examine how the rankings of thirteen forecasting models — ranging from classical methods like \texttt{ARIMA} and \texttt{ETS} to advanced deep learning models such as \texttt{NHITS} and \texttt{Informer} — are influenced by different dataset selections. We use a set of thirteen diverse benchmark datasets commonly reported in time series forecasting papers. This setup allows us to explore the impact of selective dataset reporting (cherry-picking) on model performance. Many of these models have been reported as best in class. Our goal is to determine whether the choice of datasets significantly influences these rankings and whether these models would still be considered top performers across different dataset scenarios.

We focus on three key research questions:

\begin{itemize}
    \item \textbf{Q1}: How does the selection of datasets impact the overall ranking of time series forecasting models?
    \item \textbf{Q2}: How does cherry-picking specific datasets influence the perceived performance of models?
    \item \textbf{Q3}: How many models could be reported as top performers using a cherry-picked subset of datasets?
\end{itemize}

\subsection{Datasets}\label{sec:data}

We use a diverse set of benchmark datasets covering various sampling frequencies, domains, and applications. They are summarized in Table\ref{tab:data}.

\subsection{Methods}\label{sec:methods}

The experiments include thirteen forecasting approaches, encompassing both classical and advanced deep learning methods. 

We start by introducing the classical approaches:

\begin{itemize}
    \item \texttt{SNaive}: This method forecasts future values based on the last observed value from the same season in previous cycles. 

    \item \texttt{RWD (Random Walk With Drift)} \cite{hyndman2018forecasting}: This method extends the naive forecasting approach by adding a drift component, which represents the average change observed in the historical data.
    
    \item \texttt{ETS} \cite{hyndman2008forecasting}: This approach models time series data by accounting for level, trend, and seasonality components.
    
    \item \texttt{ARIMA} \cite{hyndman2008automatic}: A widely used statistical method for time series forecasting that models data using its own past values and past forecast errors.
    
    \item \texttt{Theta} \cite{assimakopoulos2000theta}: This method decomposes a time series into two or more Theta lines, each processed separately before being recombined.
    
    \item \texttt{SES (Simple Exponential Smoothing)} \cite{hyndman2008forecasting}: This method forecasts future values by exponentially weighting past observations, giving more weight to recent data points.
    
    \item \texttt{Croston}: The method is specifically designed for intermittent demand series.
\end{itemize}

The study also incorporates six deep learning architectures. These models are noted for their advanced capabilities in capturing complex patterns in time series data:

\begin{itemize}
    \item \texttt{RNN}  \cite{ELMAN1990179}: \texttt{RNN}s are a class of neural networks that can model sequential data by maintaining a hidden state that captures information from previous time steps.
    
    \item \texttt{TCN} \cite{oord2016wavenetgenerativemodelraw}: \texttt{TCN}s are specialized for time series data, utilizing convolutional layers with dilated convolutions to capture long-range dependencies.
    
    \item \texttt{DeepAR} \cite{salinas2019deeparprobabilisticforecastingautoregressive}: This method combines autoregressive models with deep learning to handle complex time series data.
    
    \item \texttt{NHITS} \cite{challu2023nhits}:  \texttt{NHITS} builds upon NBEATS by using hierarchical interpolation and multi-rate input processing. 
    
    \item \texttt{TiDE} \cite{das2024longtermforecastingtidetimeseries}: The \texttt{TiDE} model is a Multi-layer Perceptron (MLP) based encoder-decoder designed for long-term time series forecasting. 

    \item \texttt{Informer} \cite{zhou2021informerefficienttransformerlong}:  \texttt{Informer} is a transformer-based model tailored for long sequence time-series forecasting. 
\end{itemize}

\section{Results and Discussion}\label{sec:results}

In this section, we present the results of our analysis on the impact of cherry-picking datasets in the evaluation of time series forecasting models.

We start by answering \textbf{Q1}. The selection of datasets has a significant impact on the overall ranking of time series forecasting models. Our findings indicate that while some models demonstrate robustness across a wide range of datasets, most are very sensitive to the specific datasets used in their evaluation.

\begin{figure}[H]
\centering
\includegraphics[width=1\columnwidth]{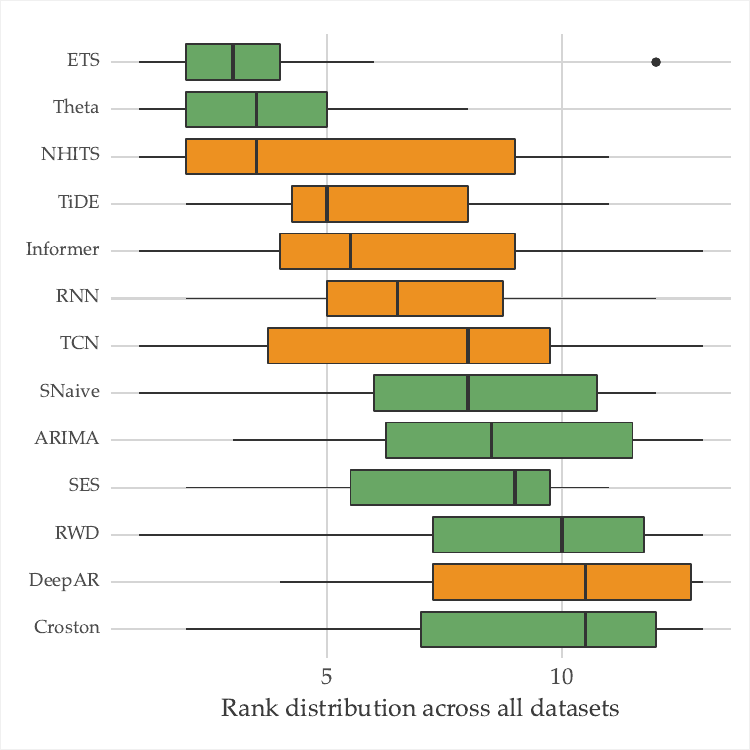}
\caption{Rank distribution of various forecasting models across all datasets. Models are organized vertically, with green bars representing classical methods and orange bars representing deep learning models..}
\label{fig:overall_rank}
\end{figure}

We examine the overall rank distribution of the models across all datasets. Figure~\ref{fig:overall_rank} presents a box plot showing the rank distribution for each model when evaluated across the entire dataset collection. This figure serves as a baseline for understanding how models perform without any cherry-picking.

\begin{figure*}[h!]
\centering
\includegraphics[width=0.33\textwidth]{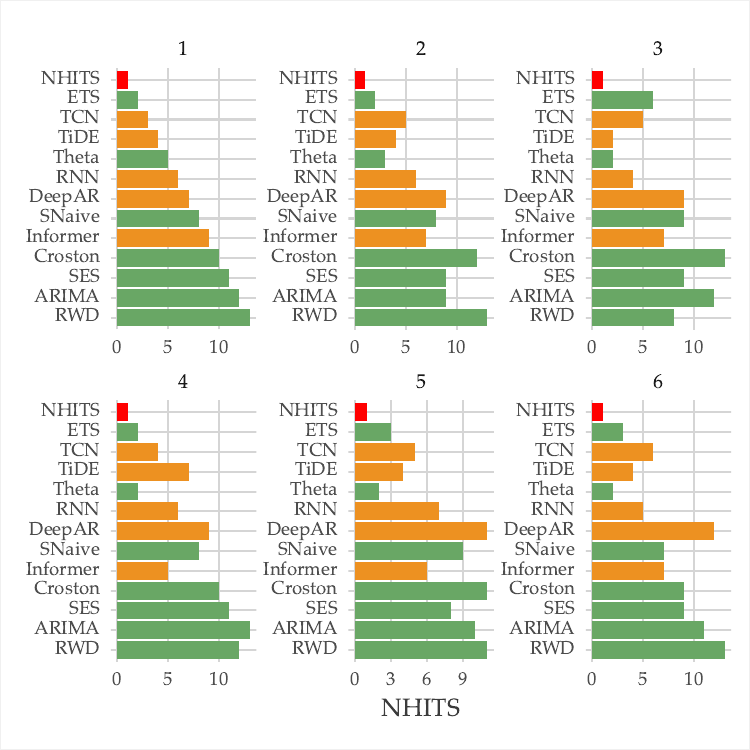}
\includegraphics[width=0.33\textwidth]{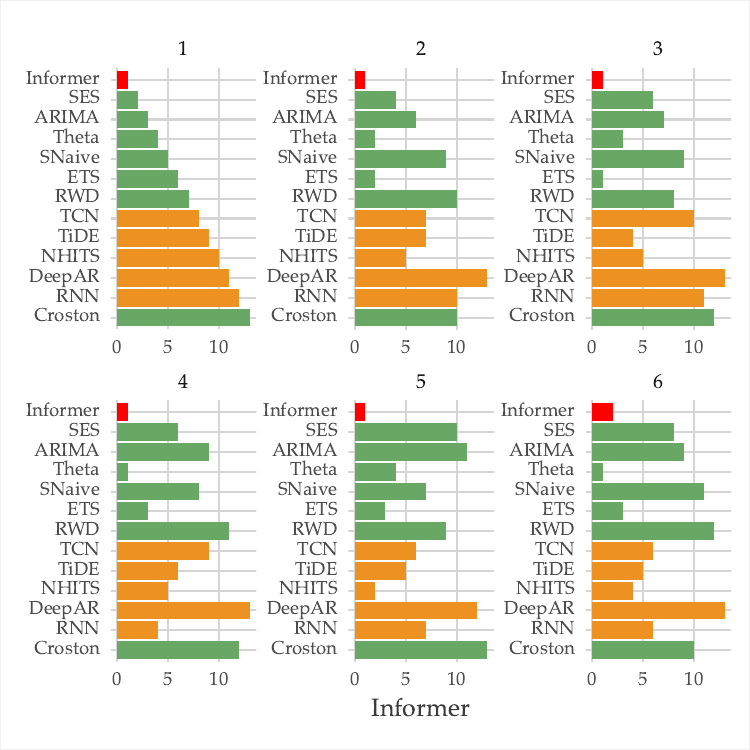}
\includegraphics[width=0.33\textwidth]{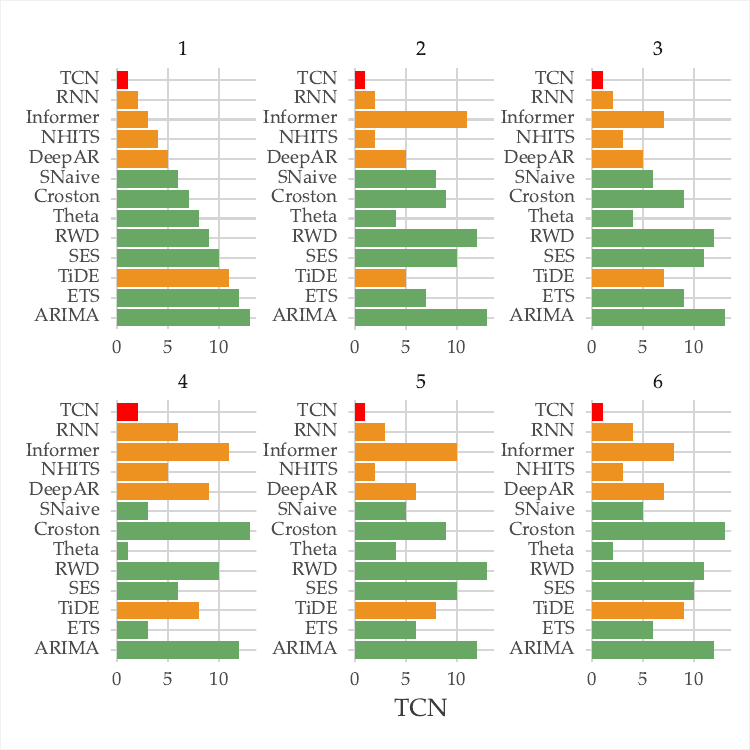}
\caption{Impact of cherry-picking on the rankings of NHITS (left), Informer (center), and TCN (right). Each subfigure (1 through 6) represents the model rankings based on cherry-picked subsets. The red bars indicate the model that we are cherry-picking for.}
\label{fig:cherrypicking}
\end{figure*}

While both \texttt{NHITS} and \texttt{ETS} models show the best median ranking, the \texttt{ETS} model demonstrates greater robustness, as indicated by its narrower interquartile range (IQR). Despite the strong median performance of \texttt{NHITS}, its ranking can drop significantly, reaching as low as rank 10 on some datasets. Conversely, \texttt{ETS} shows a relatively stable performance, though it does have one outlier where it ranks 12th. This variability, even in the best models, shows the potential for bias in experimental results if such extreme cases are included.

Other models, such as  \texttt{TCN} and \texttt{Informer}, exhibit much larger variances, with rankings ranging from 1 to 13, depending on the dataset. Among the models evaluated, \texttt{DeepAR} consistently performs the worst, showing a wide range of rankings with consistently low performance across the board. This wide variance indicates that \texttt{TCN}, \texttt{Informer}, and \texttt{DeepAR} are particularly sensitive to the specific datasets used in their evaluation, making them even more susceptible to the influence of dataset characteristics.

Regarding \textbf{Q2}, our findings demonstrate that cherry-picking specific datasets can significantly inflate the perceived performance of time series forecasting models. By selectively choosing datasets, models like  \texttt{Informer} and  \texttt{TCN} can be made to appear as top performers, even when their overall robustness may not fully justify such rankings.

Figure~\ref{fig:cherrypicking} illustrates the impact of selectively choosing datasets on the rankings of three models: \texttt{NHITS}, \texttt{Informer}, and \texttt{TCN}. The figure shows how the rank of each model changes as the number of cherry-picked datasets increases from 1 to 6.

We observe that  \texttt{Informer} and \texttt{TCN} could be reported as the best models in an experimental setup using up to 5 or even 6 datasets in the TCN case. This selective reporting would portray these models as highly effective, exaggerating their generalizability and robustness.

\texttt{NHITS}, which has demonstrated significantly more robustness compared to other models, could be reported as a top model across all 6 datasets. Note how similar it is to how  \texttt{Informer} and  \texttt{TCN} might be reported in cherry-picked setups. Also, note that in those cases, both  \texttt{Informer} and  \texttt{TCN} rank higher than \texttt{NHITS} for all \(n\).

Answering \textbf{Q3}, our analysis reveals that cherry-picking datasets can significantly skew the perceived performance of forecasting models. It makes it possible to present a large proportion of models as top performers. 

\begin{figure}[H]
\centering
\includegraphics[width=1\columnwidth]{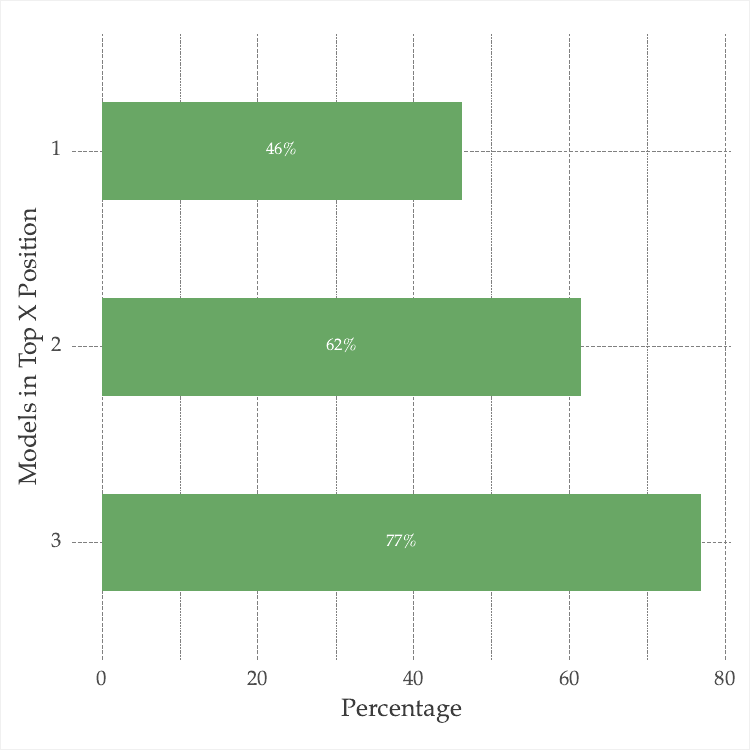}
\caption{Percentage of models that could be reported as top 1, 2, and 3 performers based on an experimental setup of 4 datasets.}
\label{fig:top_n_4}
\end{figure}

Figure~\ref{fig:top_n_4} shows that when reporting with just 4 cherry-picked datasets, you could make 46\% of the models in our experimental setup appear as the best model. Additionally, 77\% of the models could be reported as ranking within the top 3. 

\begin{figure}[H]
\centering
\includegraphics[width=1\columnwidth]{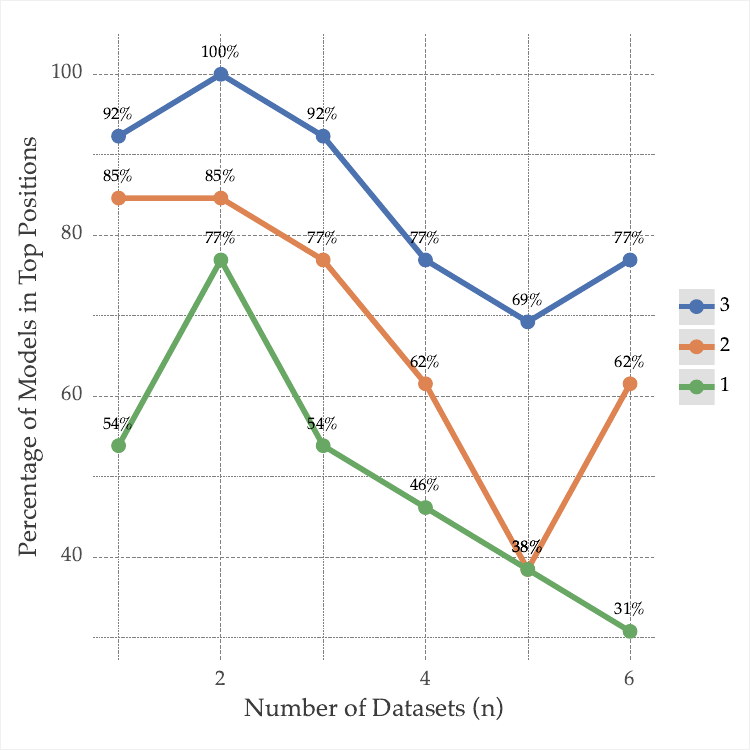}
\caption{Breakdown of the percentages for top 1, 2, and 3 positions across different numbers of datasets.}
\label{fig:top_n}
\end{figure}

In Figure~\ref{fig:top_n}, we see how easy it is to present models as top performers solely through cherry-picking. For instance, with an experimental setup of 3 datasets (commonly found), more than 54\% of models could be reported as the top 1 performer and 92\% as the top 3. Even with 6 datasets, it is still possible to report 77\% of models as ranking within the top 3 positions. This conveys the persistence of potential bias, even as the dataset size increases.

\subsection{Conclusions}\label{sec:conclusions}

The main conclusion of this work is that selectively choosing datasets can significantly distort the perceived performance of forecasting models, leading to biased reporting. This practice can obscure the true capabilities, making it appear more effective than it may be across diverse applications.

Our findings highlight the need for rigorous, unbiased, and standardized evaluation methodologies in time series forecasting. Adopting such practices is essential to ensure that models are robust and reliable across a wide range of real-world scenarios. Additionally, it reduces the risk of overestimating their effectiveness based on selectively reported outcomes.

\section{Acknowledgments}
This work was partially funded by projects AISym4Med (101095387) supported by Horizon Europe Cluster 1: Health, ConnectedHealth (n.º 46858), supported by Competitiveness and Internationalisation Operational Programme (POCI) and Lisbon Regional Operational Programme (LISBOA 2020), under the PORTUGAL 2020 Partnership Agreement, through the European Regional Development Fund(ERDF) and Agenda “Center for Responsible AI”, nr. C645008882-00000055, investment project nr. 62, financed by the Recovery and Resilience Plan (PRR) and by European Union - NextGeneration EU, and also by FCT plurianual funding for 2020-2023 of LIACC (UIDB/00027/2020 UIDP/00027/2020);

\bibliography{aaai25}

\end{document}